\def\year{2019}\relax
\documentclass[letterpaper]{article} 
\usepackage{papers}  
\usepackage{times}  
\usepackage{helvet}  
\usepackage{courier}  
\usepackage{url}  
\usepackage{graphicx}  
\usepackage{bm}
\usepackage{amsmath}
\begin{document}
%
\title{Fast Neural Chinese Word Segmentation for Long Sentences}
\author{Sufeng Duan, Jiangtong Li, Hai Zhao\thanks{Corresponding author. This paper was partially supported by National Key Research and Development Program of China (No. 2017YFB0304100),
Key Projects of National Natural Science Foundation of China (U1836222 and 61733011),
Key Project of National Society Science Foundation of China (No. 15-ZDA041),
The Art and Science Interdisciplinary Funds of Shanghai Jiao Tong University (No. 14JCRZ04).}\\
	Department of Computer Science and Engineering, Shanghai Jiao Tong University \\
	Key Laboratory of Shanghai Education Commission for Intelligent Interaction \\
    and Cognitive Engineering, Shanghai Jiao Tong University, Shanghai, 200240, China\\
    \{\tt1140339019dsf, keep\_moving-lee\}@sjtu.edu.cn, zhaohai@cs.sjtu.edu.cn\\
}
\maketitle
\begin{abstract}
Rapidly developed neural models have achieved competitive performance in Chinese word segmentation (CWS) as their traditional counterparts. However, most of methods encounter the computational inefficiency especially for long sentences because of the increasing model complexity and slower decoders. This paper presents a simple neural segmenter which directly labels the gap existence between adjacent characters to alleviate the existing drawback. Our segmenter is fully end-to-end and capable of performing segmentation very fast. We also show a performance difference with different tag sets. The experiments show that our segmenter can provide comparable performance with state-of-the-art.
\end{abstract}

\section{Introduction}
\noindent Word segmentation is the process of dividing text into words. Different from alphabetical languages like English whose words are separated by space, most of Asian languages like Chinese and Japanese have no word boundaries, for whom word segmentation is a fundamental step for most language processing tasks.

Since \cite{xue2003chinese}, most methods treat supervised Chinese word segmentation (CWS) task as a sequence labeling task with character position tags. Some CWS models such as Maximum Entropy (ME)\cite{xue2003chinese} and Conditional Random Fields (CRF)\cite{lafferty2001conditional,peng2004chinese} are applied to such a labeling formalization while performance of these models depend on handcrafted features heavily.

With the development of neural network and machine learning, a number of researchers have used neural network to improve the performance of CWS and minimize the efforts in feature engineering. \cite{zheng2013deep} adapted the sliding-window based sequence labeling with character embeddings as input instead of characters. \cite{pei2014max} introduced a model exploiting tag embeddings and tensor-based transformation. \cite{chen2015gated} introduced a gated recursive neural network (GRNN). \cite{chen2015long} proposed a model with the long short-term memory (LSTM) neural network to capture long-distance context. \cite{xu2016dependency} integrated GRNN and LSTM for deeper feature extraction. \cite{LiuCGQL16} used semi-CRF with neural network. \cite{aclCaiZ16} proposed a framework for representing words candidates from their member characters with LSTM for further performance improvement.

It is a common practice for a neural segmentation model to take up days for training. To get a better performance, the neural model can be very complex with many neural layers and functions which makes training slowly. Neural segmenter also works much slower than traditional ones \cite{CaiZZXWH17}. Except for models with greedy decoder, most of existing models may fall into two types of decoding time complexities. For those sequence labeling models, either traditional or neural ones, only if they consist a first-order Markov mechanism, they will all share the same time complexity, $O$($Mnt^2$), where $n$ is the number of characters in a sentence, $t$ is the number of labeling tags ($t$=2 for B, I and $t$=4 for B, M, E, S tagging schemes) and $M$ is the model complexity for computing local probability distribution. For those models with beam search decoder, they also have a similar time complexity, $O$($Mnb^2$), where $b$ is the beam width (popularly set to 10 for most known models). Considering that usually it is neural models that have to use a beam search decoder and have much higher model complexity than traditional ones (i.e., $M$ is much larger), neural segmenters developed in recent years are actually much more inefficient than those early traditional ones.

According to the above analysis, the longer the sentence is, the slower the segmentation is. Not only that, it is also hard for a model to capture long distance dependence. Actually, most of existing models cannot deal with long sentences as good as short ones, which makes the long sentence segmentation both slow and inaccurate.

In this paper, we propose a fast neural segmenter for CWS. The segmenter is based on a latest developed attention mechanism, Biaffinal attention, which generally aims to lead the model to pay specific attention to different part of a specific task, since \cite{Bahdanau15} first introduced the mechanism into the neural machine translation. Unlike other labeling based CWS models which need to predict the label of characters, our segmenter straightforwardly models the sentences and directly predicts the segmentation gap between two adjacent characters. Our model includes an encoder and a gap scorer, which is capable of scoring the gap existence and can be trained and tested end-to-end. There is no decoder in our model that means our model can predict fast and save time by skipping the decoding process. Therefore, our model can train and predict very fast with better long sentence performance than short ones.

\begin{figure*}[!htb]
	\centering
	\includegraphics[scale=0.7]{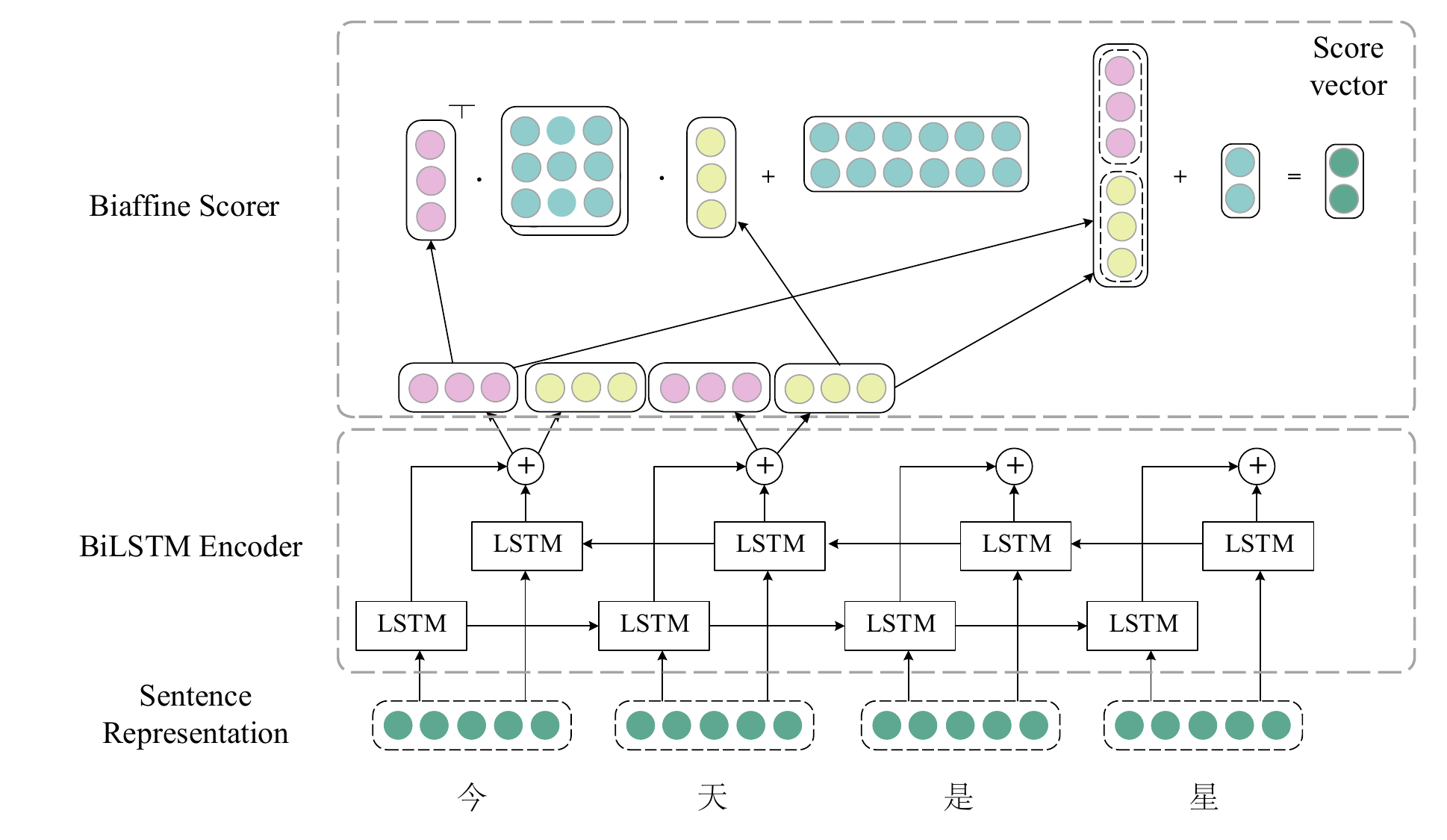}
	\caption{An overview of our model.} \label{biaff_model}
\end{figure*}

Based on the idea of attention mechanism, we also develop several models for different tag sets such as $\{B,E,M,S\}$. In the experiments we compared these models and found difference of performances between our segmenter and traditional tag based segmenter.


The remainder of the paper is organized as follows: Section 2 introduces the related work. Section 3 describes our fast neural segmenter. Section 4 presents our experiments and Section 5 gives a analysis for the result of experiments.

\section{Related Work}
In this section, we review the preview works about Chinese Word Segmentation and biaffinal attention.

Chinese Word Segmentation is a fundamental step for most Chinese natural language processing tasks and has been well studied for decades. Xue \shortcite{xue2003chinese} was the first to formalize CWS tasks as a character-based tagged problem. Peng et al. \shortcite{peng2004chinese} proposed a CRF based model to solve CWS tasks. Following these achievements, some Chinese segmenter \cite{tseng2005conditional,zhao2006effective,zhao2008unsupervised,zhao2010unified,sun2012fast,zhang2013exploring} were proposed and got better performances. The method that transforms CWS into sequence labeling problem has been used in some neural models \cite{zheng2013deep,pei2014max,chen2015gated,chen2015long}. \cite{huang2006essential,sun2010word,wang2014two} studied CWS model with both character-based and word-based segmenters. \cite{aclCaiZ16} proposed a segmentation model which replaced a fixed sized sliding window with a feature window to cover complete input and segmentation history. Following the work\cite{aclCaiZ16}, Cai et al.\shortcite{CaiZZXWH17} present a greedy neural word segmenter with balanced word and character embeding input that performing segmentation much faster.

Traditionally, labeling task based CWS can find gap by labeling characters in a sentence. Based on this idea, Zhao et al.\shortcite{zhao2006effective} compared 2-tags, 4-tags, 5-tags and 6 tags model. While CWS can also predict the gap directly. Huang et al.\shortcite{huang2007rethinking} introduced a method that labeling gap directly instead of the character.

Our work has used biaffinal attentional mechanism. Bahdnau et al.\shortcite{Bahdanau15} introduced the traditional attentional mechanism in the neural machine translation(NMT). Dozat et al.\shortcite{dozat2017deep} proposed a mechanism that use biaffine attention instead of bilinear attention or MLP-based attention. In this paper, the part of biaffine attention can evaluate score between different words.

\section{Models}
As shown in Figure \ref{biaff_model}, our model is divided into two modules, (1) a bidirectional LSTM (BiLSTM) encoder that takes each character embedding $e_i$ of the given sentence as input and generates dense vectors for all the characters, (2) a biaffine attentional scorer which takes the hidden vectors for all given character pairs as the input and predict a label score vector.

\subsection{Bidirectional LSTM Encoder}
\paragraph{Character Representation}
Given a sentence, following \cite{Bengio2006A} we use a lookup table to transform this sequence of characters into a sequence of character embedding $\bm{S_e} = \{\bm{e}_1, \bm{e}_2, ... , \bm{e}_n\}$.

\paragraph{Encoder}
BiLSTM is adopted for our sentence encoder. By incorporating a stack of two distinct LSTMs, BiLSTM processes an input sequence both forwardly and backwardly, then combines the outputs of two LSTMs as the representation of a character.

Given a sequence of character embedding $\bm{S_e} = \{\bm{e}_1, \bm{e}_2, ... , \bm{e}_n\}$ as input, the $\bm{i}$-th element $\bm{g_i}$ is encoded as follows:

\begin{center}
$\bm{g}^\mathcal{F}_i = LSTM^\mathcal{F}\left(\bm{r}_i, \bm{g}^\mathcal{F}_{i-1}\right),$\\
$\bm{g}^\mathcal{B}_i = LSTM^\mathcal{B}\left(\bm{r}_i, \bm{g}^\mathcal{B}_{i+1}\right),$\\
$\bm{g}_i   = \bm{g}^\mathcal{F}_i \oplus \bm{g}^\mathcal{B}_i,$
\end{center}
where $LSTM^\mathcal{F}$ and $LSTM^\mathcal{B}$ respectively denote the forward and backward LSTM transformation, $\bm{g}^\mathcal{F}_i$ and $\bm{g}^\mathcal{B}_i$ are the hidden state vectors of the forward and backward LSTMs respectively and  $\oplus$ denotes the concatenation operation.

\subsection{Biaffine Attentional Scorer}

Typically, to label the gap existence between two adjacent characters, a segmentation scorer is employed on the top of BiLSTM encoder, which is implemented as biaffine attention introduced by \cite{dozat2017deep}, for compared to its counterparts such as bilinear or MLP-based attention, the biaffine attention is more effectively capable of measuring the relationship between two elementary units. Note that biaffine attention is a natural extension of bilinear attention \cite{LuongPM15} which is widely used in neural machine translation (NMT).

\paragraph{Affine Transformation}
For a task like CWS, the scorer is supposed to distinguish the gap existence between two adjacent characters. To this end, we perform two distinct affine transformations on the hidden state $\bm{g}_i$, mapping it to vectors with smaller dimensionality:

\begin{center}
$\bm{h}_i^{(front)}  = \bm{W}^{(front)}\bm{g}_i + \bm{b}^{(front)},$\\
$\bm{h}_i^{(rear)} = \bm{W}^{(rear)}\bm{g}_i + \bm{b}^{(rear)},$
\end{center}
 where $\bm{h}_i^{(front)}$ and $\bm{h}_i^{(rear)}$ are the hidden states representation respectively for the front and the rear characters.

By make such transformations over output from encoder, the scorer can benefit from deeper feature extraction. Both features of adjacent characters are learned by the same LSTMs, the scorer can get features composed from both recurrent states together with reduced dimensionality. This model also map the front character and the rear character into two distinct spaces which can help the scorer plays in different context and simple.

\paragraph{Biaffine Scoring}

Traditionally, bilinear transformation is a good tool to judge the relationship between two objective vectors. In bilinear transformation, given a target recurrent output vector $h_i^{(t)}$ and a source recurrent output vector $h_j^{(s)}$, a bilinear transformation calculates a score $s_{ij}$ for the alignment:
\begin{center}
$s_{ij} = \bm{h}_i^{\top (t)} \bm{W} \bm{h}_j^{(s)},$
\end{center}

In a traditional classification task, the distribution of different classes is often uneven so that the output layer of the model normally includes a fixed bias term designed to capture the prior probability $P(y_i = c)$ of each class, with the rest of the model focusing on learning the likelihood of each class given the data $P(y_i = c|x_i )$. In order to alleviate the burden of the fixed bias term and capture the prior probability dynamically, bias terms are introduced into the bilinear attention resulting in a biaffine transformation.

In CWS, the distribution of the gap existence is similarly uneven, so directly applying the primitive form of bilinear attention would fail to capture the prior probability $P(y_i=c_k)$ for each class. Thus, the biaffine attention introduced in our model would be extremely helpful for gap existence prediction.

\begin{displaymath}
\begin{aligned}
\bm{s}_{i} = \ &\bm{h}_i^{\top (front)} \bm{W}^{(gap)} \bm{h}_{i+1}^{(latter)} \\ +\ & \bm{U}^{(gap)}\left( \bm{h}_i^{(front)}\oplus \bm{h}_{i+1}^{(latter)} \right) \\ +\ &\bm{b}^{(gap)},
\end{aligned}
\end{displaymath}
where $\bm{W}^{(gap)}$, $\bm{U}^{(gap)}$ and $\bm{b}^{(gap)}$ will be updated during the learning process.

Given a sentence of length $n$, for every adjacent character pairs, the scorer outputs a score vector $\bm{S} = \{\bm{s}_1, \bm{s}_2, ... , \bm{s}_{n}\}$. Then our model selects as its output the label with the highest score from each score vector: $y_{i}=\mathop{\arg\max} \bm{S}[i]$, where $\bm{S}[i]$ denotes the score of the $i$-th gap existence.
\subsection{Tag Sets}
\begin{table}[!htb]
	\centering
	
	\setlength{\tabcolsep}{1.2mm}{
		\begin{tabular}{l|cc|cc}
			\hline
			{\textbf{Tag set}} &{\textbf{Tags}} &{\textbf{Words in tagging}} \\
			\hline
			\textbf{2-tag} & B,E & B,BE,BEE,... \\
			\textbf{4-tag} & B,M,E,S &S,BE,BME,BMME,... \\
			\hline
		\end{tabular}
	}
	\caption{Definitions of $\{B,E\}$ and $\{B,E,M,S\}$ tag set}
	\label{tag_set}
\end{table}

Tradiationlly speaking, most of previous segmenters have to label every character and use a decoder to find the gap in a sentence. However, how to select an effective tag set for a segmentation task is an interesting and important problem. For a basic segmenter, there are two major kinds of schemes known as $\{B,E\}$ and $\{B,E,M,S\}$. The detailed information is in Table \ref{tag_set}.

Different from other segmenter, our segmenter directly predicts the gaps in a sentence which leaves out the decoding step. The idea of our segmenter is that the gap is actually a special relationship of a pair of adjacent characters. Introduced by \cite{dozat2017deep}, biaffine attention is a tool to score every pairs of element in a sequence. Our segmenter use the biaffine transformer based classifier to predict the gap existence directly, which is the simple and doesn't need a decoder. However, inspired by the traditional method, we also find a way to combine the character labeling method and the gap labeling method. In this work, we present three different methods which is gap labeling based and using different label to combine the character labeling together.

To further explore different method in labeling the gap existence, we combine the $\{B,E\}$ and $\{B,E,M,S\}$ labeling schemes into our model. In the combined model, the label of one gap can be defined by labels of the two adjacent characters. The set of labels with $\{B,E\}$ tag set can be $\{BB, BE, EE, EB\}$. Similar as $\{B,E\}$ tag set, the set of labels with $\{B,E,M,S\}$ tag set can be $\{BM, BE, ME, ES, SB, SS, MM, EB\}$. The conflict may exist similar as traditional segmenter with $\{B,E\}$ and $\{B,E,M,S\}$. So the segmenter needs a decoder to guarantee the validity of a sentence.

For $\{B,E\}$ and $\{B,E,M,S\}$ model, every label of gap has only two candidacy (as Tables \ref{tag_set_trans_2_tag} and \ref{tag_set_trans_4_tag} shows) that limit the time complexity of decoder. As a result the time complexity of decoder only depend on the beam-size and the length of sentence.
\begin{figure}
	\centering
	\includegraphics[scale=0.3]{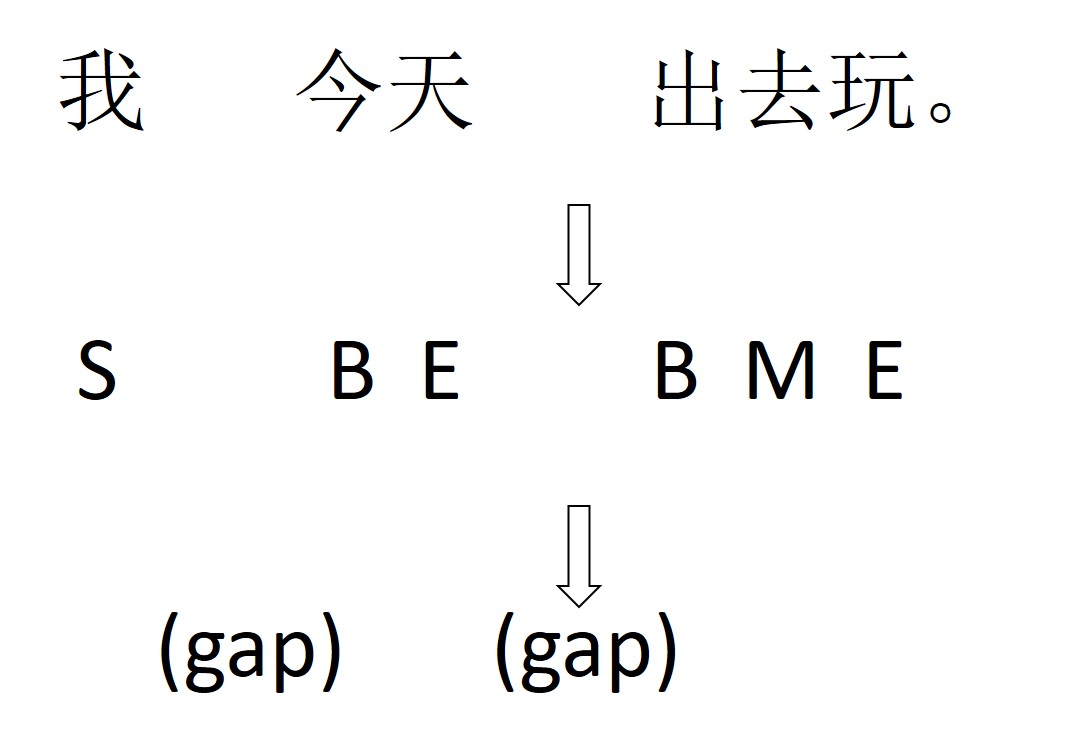}
	\caption{The step of traditional segmenter works.}
    \label{traditional}
\end{figure}
\begin{figure}
	\centering
	\includegraphics[scale=0.3]{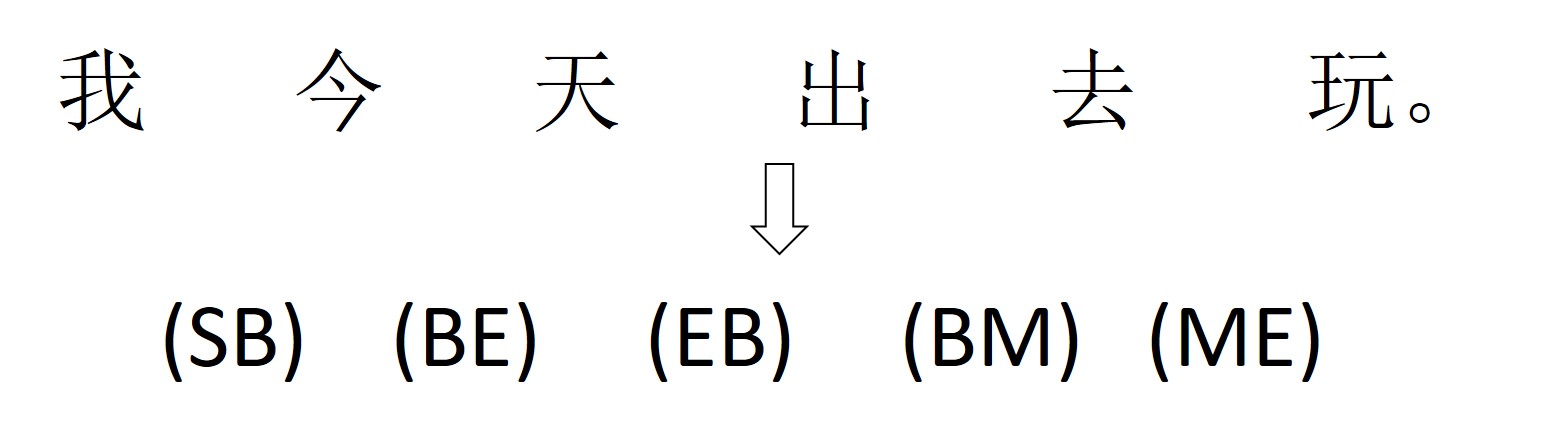}
	\caption{The step of our BEMS segmenter works.}
    \label{traditional}
\end{figure}
\begin{table}[!htb]
	\centering
	\large
	\setlength{\tabcolsep}{1.2mm}{
		\begin{tabular}{l|cc|cc}
			\hline
			{\textbf{Tags}} &{\textbf{Words in tagging}} \\
			\hline
			 BE & EB,EE \\
			 BB & BB,BE \\
            EB & BB,BE \\
            EE & EB,EE \\
			\hline
		\end{tabular}
	}
	\caption{Transition of $\{B,E\}$ tag set}
	\label{tag_set_trans_2_tag}
\end{table}
\begin{table}[!htb]
	\centering
	\large
	\setlength{\tabcolsep}{1.2mm}{
		\begin{tabular}{l|cc|cc}
			\hline
			{\textbf{Tags}} &{\textbf{Words in tagging}} \\
			\hline
			 BE & EB,ES \\
			 BM & ME,MM \\
            EB & BE,BM \\
            ES & SB,SS \\
             SS & SS,SB \\
			 SB & BE,BM \\
            ME & EB,ES \\
            MM & MM,ME \\
			\hline
		\end{tabular}
	}
	\caption{Transition of $\{B,E,M,S\}$ tag set}
	\label{tag_set_trans_4_tag}
\end{table}

Here we have three different methods to label the gap and experiment will tell us which is the next direction for CWS.

\section{Experiments}

\subsection{Datasets and Settings}
\begin{table}[!htb]
    \centering
    \large
    \setlength{\tabcolsep}{0.9mm}{
    \begin{tabular}{l|cc|cc}
    \hline
     &\multicolumn{2}{|c|}{\textbf{AS}} &\multicolumn{2}{|c}{\textbf{MSR}}  \\
     & Train& Test&Train&Test\\
     \hline
     \textbf{\#sentences} &709K & 14K & 87K & 4K  \\
     \textbf{\#words} & 4741K & 108K & 2,368K & 107K \\
     \textbf{\#characters} & 8368K & 198K & 3,981K & 181K\\
    \hline
    \end{tabular}
    }
    \caption{Data statistics}
    \label{datastatis}
\end{table}

\begin{table}[!htb]
	\centering
	\large
	\setlength{\tabcolsep}{1.2mm}{
		\begin{tabular}{l|cc|cc}
			\hline
			&\multicolumn{2}{|c|}{\textbf{AS}} &\multicolumn{2}{|c}{\textbf{MSR}}  \\
			Length & \#sen(\%) & \#char(\%) & \#sen(\%) & \#char(\%) \\
			\hline
			\textbf{0-30} & 0.875 & 0.705 & 0.270 & 0.126 \\
			\textbf{31-60} & 0.135 & 0.263 & 0.486 & 0.434 \\
			\textbf{61-90} & 0.007 & 0.045 & 0.175 & 0.266 \\
			\textbf{91-120} & 0.001 & 0.004 & 0.047 & 0.100 \\
			\textbf{121-inf} & 0.001 & 0.002 & 0.022 & 0.072 \\
			\hline
		\end{tabular}
	}
	\caption{Sentence length statistics}
	\label{lenghtstatics}
\end{table}

\begin{table}[!htb]
	\centering
	
	\setlength{\tabcolsep}{1.2mm}{
		\begin{tabular}{l|l|l|l}
            \hline
            {Parameters} & {$\{0,1\}$} & {$\{B,E\}$} & {$\{B,M,E,S\}$} \\
			\hline
			{Character embedding size} & {300}& {300}& {300} \\
			{Depth of LSTM layers} & {3} & {3} & {3} \\
			{Hidden state size} & {300} & {300}& {300}\\
			{Biaffine input size} & {300} & {300}& {300}\\
			{Learning rate} & 0.001 & 0.0012 & 0.002\\
			{Dropout probability} & 0.6 & 0.39 & 0.45\\
			\hline
		\end{tabular}
	}
	\caption{Hyperparameter settings}
	\label{model_set}
\end{table}
We evaluate our model on two popular benchmark datasets, namely AS and MSR from the second international Chinese word segmentation bakeoff \cite{Emerson2005}, which is the biggest two of all the benchmarks. The data statistics are in Table \ref{datastatis}.

To compare performance of different tag sets, we also evaluate the model with $\{0,1\}$ tag, $\{B,E\}$ tag set and $\{B,E,M,S\}$ tag set in AS and MSR datasets. For $\{B,E\}$ tag set and $\{B,E,M,S\}$ tag set, the model uses beam-search decoder to get the result of sentences, and the beam size of decoder is 10. For $\{0,1\}$ tag, the model directly predicts the gap of the input.

The statistics about sentence length distribution is shown in Table \ref{lenghtstatics}. As for AS dataset, most of sentences are short due to its annotation including a manually sentence splitting preprocessing. Even so, long sentences (with more than 30 characters) carry more than $25\%$ characters of the entire dataset. As for the MSR, without manually sentence splitting, more than $70\%$ sentences are longer than 30 characters, which means that fast and accurate segmentation for long sentences is desperately in need.

In this paper, we use the same model setting as show in Table \ref{model_set}. These numbers are tuned on development sets.\footnote[1]{Following conventions, the last 10\% sentences of training	corpus are used as development set.} The character embeddings are pre-trained by word2vec \cite{Mikolov} toolkit using skip-gram on Chinese Wikipedia corpus. The learning rate and dropout probability of model is different with different tag set while other parameters are the same. During the training, we use dropout layer before every affine transformation to alleviate the overfitting problem. Our model is optimized using Adam \cite{adam2015}.

\begin{table}
	\centering
	\large
	\begin{tabular}{l|ccc}
		\hline
		AS& F-1 & Train(h) & Test(s)  \\
		\hline
		\cite{wang2014two} & 95.4&-&-\\
		\cite{ChenSQH17}  &94.5&-&-\\
		\cite{CaiZZXWH17}&95.0&60&80\\
        \hline
		our model(01) & 94.4 & 9 & 10 \\
        our model(BE) & 94.5 & 12.25 & 112\\
        our model(BEMS) & 94.8 & 16 &150\\
		\hline
	\end{tabular}
	\caption{Comparison of performance and running time}
	\label{AS}
\end{table}
\begin{table}
	\centering
	\large
	\begin{tabular}{l|ccc}
		\hline
		MSR & F-1 & Train(h) & Test(s)  \\
		\hline
		\cite{chen2015gated} &95.4&100&120\\
		\cite{chen2015long}  &95.6&117&120\\
		\cite{aclCaiZ16}  &96.5& 96 & 105\\
		\cite{CaiZZXWH17}&97.1&6&30\\
		\cite{Jianqiang} &96.6&3&28\\
        \hline
		our model(01) & 96.6 & 2.5 & 5 \\
        our model(BE) & 96.7 & 4.6 & 52 \\
        our model(BEMS) & 96.9 & 10.2 & 90\\
		\hline
	\end{tabular}
	\caption{Comparison of performance and running time}
	\label{msr}
\end{table}

\begin{table}[!htb]
	\centering
	\large
	\setlength{\tabcolsep}{0.5mm}{
		\begin{tabular}{l|cc|cc}
			\hline
			&\multicolumn{2}{|c|}{\textbf{AS}} &\multicolumn{2}{|c}{\textbf{MSR}} \\
			\cline{2-5}
			& Long(s) & Short(s) & Long(s) & Short(s) \\
			\hline
			\cite{CaiZZXWH17} & 45 & 37 & 27 & 8 \\
			Our model & 5 & 7 & 4 & 3 \\
			\hline
		\end{tabular}
	}
	\caption{Predicting time in short and long sentences\footnotemark[3]}.
	\label{lenghtresult}
\end{table}
\footnotetext[2]{More than 30 characters are regarded as long sentences.}

\subsection{Main Result \footnote[3]{We use the same hardware setting as \cite{CaiZZXWH17} in this section}}

In the experiment, we test all three different tag set in the model. Tables \ref{AS} and \ref{msr} compare our final results to prior neural models\footnote[4]{We are aware there are a lot of neural models exploiting extra resources for further performance improvement, which should belong to the open test setting defined by SIGHAN-bakeoff shared task. However, to focus on the model improvement, we principally follow the closed test setting of SIGHAN-bakeoff, which only allows training dataset is used for segmenter learning. The only exception is that our comparison additionally allows all models using standard pre-trained embedding, which has been a common practice for all neural segmentation models. To let this work have an explicit focus, we thus exclude all open test concerned only work, and also those using complicated pre-trained embeddings like \cite{Yang2017Neural} and \cite{Zhou2017Word}.}. For performance, our proposed BEMS based model is comparable to the state-of-the-art models only with slight difference and faster than methods except \cite{Zhou2017Word} and \cite{Jianqiang}. For efficiency, our 01 tag based model is much faster than state-of-the-art model in \cite{CaiZZXWH17} and \cite{Jianqiang}, which are so far the fastest reported model on either training or testing. To show the efficiency of 01 tag based model clearly, Table \ref{lenghtresult} further compares \cite{CaiZZXWH17} and our 01 tag model with long and short sentences respectively, which demonstrates our model earns much more efficiency improvement over long sentences.

\subsection{Model Analysis \footnote[5]{All the results in this section are finished in development set.}}
\begin{figure}
	\centering
	\includegraphics[scale=0.5]{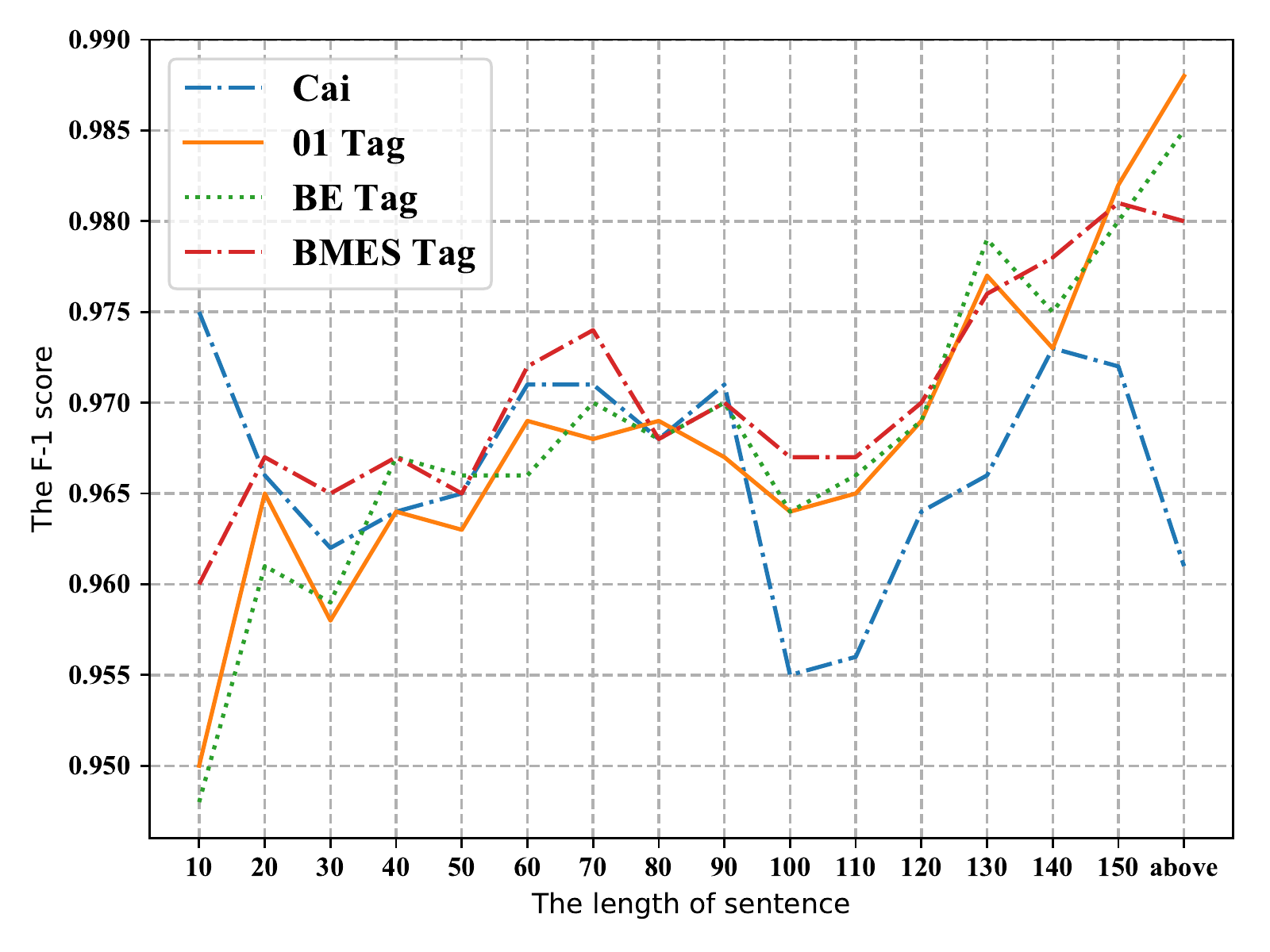}
	\caption{The comparison of our model and \cite{CaiZZXWH17} dealing with different length of sentences .} \label{length}
\end{figure}


\paragraph{Performance in Different Length}
Figure \ref{length} compares the performance of \cite{CaiZZXWH17} and our model dealing with different sized sentences in MSR. Figure shows that our model with different tags have similar trend that the longer sentence is, the better segmenter performs. Unlike the state-of-the-art model of \cite{CaiZZXWH17} which performs best in middle length, our model performs much better when handling long sentence than short ones. Namely, our model is capable of better handling long-distance dependence relationship. Note that the length of curve in our model is a surprise and maybe a counter-example to all sequence-level NLP tasks, including POS tagging, named entity recognition, and syntactic or semantic parsing, which usually show that the longer the sentence is, the poorer the processing effectiveness is.

We think the good performance in long sentence is caused by the bidirectional LSTM encoder. The longer the sentence is, the more information the sentence have. These information tells scorer how to find a gap. Our model is based on a bidirectional LSTM encoder which means the embedding of character contains information on two directions. So embedding of character can hold information of the entire sentence which may contain some potential structural and semantic information. Our scorer works on every adjacent character pairs with the information of the entire sentence which makes the performance in different length of sentence.

\subsection{Performance with Different Tag Set}


Tables \ref{AS} and \ref{msr} also compare our model with different tag sets. The results shows that model with $\{B,E,M,S\}$ is better than other two models while model with $\{B,E\}$. This results are similar as traditional segmenters\cite{zhao2006effective}. Segmenters with more tags can predict the gap type accurately and the features for different kind of gap can be learned by the model. And our model focus on the gap existence. Model with $\{0,1\}$-tag predicts the label of gap directly while the model pays attention to relationship of adjacent characters and ignores the relationship of characters in a same word which may affect the performance. Similar as $\{0,1\}$-tag model, $\{B,E\}$ model cannot get the boundary of one word which may affect the performance. Though the scorer can give a probability of gap existence using two adjacent characters only, the features for one character contain information of the entire sentence. So the segmenter can get features of other related gap.

The results also show that the model is faster with a smaller size of tag set during training and testing. More tags means more parameters model needs to optimize on training which makes $\{0,1\}$-tag model faster than others.

Model with $\{0,1\}$-tag predicts the label of gap directly without any conflict, and the original result model output is legal. So model with $\{0,1\}$-tag needs no decoder which makes it faster than others on testing. For $\{B,E\}$ and $\{B,E,M,S\}$ model, the time complexity of beam-search decoders are the same because one label of gap has only two candidacy (as Tables \ref{tag_set_trans_2_tag} and \ref{tag_set_trans_4_tag} shows) which limits the time complexity of decoder. So the difference of $\{B,E\}$ and $\{B,E,M,S\}$ model is cased by the complexity of model.

\subsection{Moderate Performance Improvement}
Performance of some segmenters drop when the length of sentence is large. The good performance in long sentence of our model can be used to improve other segmenters. One simple method is that to use our model to predict long sentences if the length of sentences is larger than threshold.

In the experiment, we use this method to improve model introduced by \cite{CaiZZXWH17} with our model. Figure \ref{length} shows that performance of model introduced by \cite{CaiZZXWH17} drop when length of sentences is large than 90. And we set the threshold to 90. When the length of sentence is larger than 90, the segmenter will select result of our model. The dataset is MSR.

\begin{table}
	\centering
	\large
	\begin{tabular}{l|c}
\hline
model& F-1 \\
		\hline
		 \cite{CaiZZXWH17} &  96.8\\
		\hline
		with our model(01) & 96.9  \\
        with our model(BE) & 96.9   \\
        with our model(BEMS) & 96.9 \\
		\hline
	\end{tabular}
	\caption{Performance improvement}
	\label{msrour}
\end{table}
 Table \ref{msrour} shows that the segmenter improved by $\{0,1\}$,$\{B,E\}$ and $\{B,E,M,S\}$ model have a moderate performance improvement compared with original segmenter. In this method, the original segmenter is independent of $\{0,1\}$,$\{B,E\}$ and $\{B,E,M,S\}$ model which means this method is more flexible than one segmenter.

\section{Conclusion}
This paper reports a long sentence oriented neural segmenter which straightforwardly models Chinese word segmentation as gap decision according to biaffine scoring over BiLSTM encoder. Our model can be trained and tested end-to-end with a simplified model architecture. Our model can be trained and predicted fast without decoder. We also designed two other gap types for the model. The evaluation on benchmark shows that our model extraordinarily performs with a segmentation style that the longer, the better, on both performance and efficiency.

\bibliography{papers}
\bibliographystyle{paperbst}

\end{document}